\documentclass{article}
\usepackage{spconf,amsmath,graphicx}
\usepackage{graphicx}
\usepackage{amsmath}
\usepackage{amssymb}
\usepackage{booktabs}
\usepackage{algorithm}
\usepackage{algorithmic}
\usepackage{multirow}
\usepackage{soul}
\usepackage{xcolor}
\usepackage{subcaption}
\usepackage{newfloat}
\usepackage{listings}

\usepackage{array}
\newcolumntype{P}[1]{>{\centering\arraybackslash}p{#1}}%
\usepackage{hyperref}

\newcommand{\tsc}[1]{\textsuperscript{#1}}

\usepackage{accents}
\newcommand\thickbar[1]{\accentset{\rule{.4em}{.8pt}}{#1}}

\title{\texttt{WAVER}:
\texttt{\underline{W}}riting-style
\texttt{\underline{A}}gnostic Text-\texttt{\underline{V}}id\texttt{\underline{E}}o \texttt{\underline{R}}etrieval via Distilling Vision-Language Models Through Open-Vocabulary Knowledge}

\name{
  \hspace{0.1em} Huy Le\tsc{1,2},
  \hspace{0.1em} Tung Kieu\tsc{3},
  \hspace{0.1em} Anh Nguyen\tsc{4},
  \hspace{0.1em} Ngan Le\tsc{5}
}

\address{\tsc{1} FPT Software AI Center, \tsc{2} International University, VNU-HCM, \tsc{3} RMIT University, Vietnam \\ \tsc{4} University of Liverpool, UK \hspace{0.2cm} \tsc{5} University of Arkansas, USA\\}

\begin{document}
\setlength{\abovedisplayskip}{0.5pt}
\setlength{\belowdisplayskip}{1pt}

\maketitle

\begin{abstract}
Text-video retrieval, a prominent sub-field within the domain of multimodal information retrieval, has witnessed remarkable growth in recent years. However, existing methods assume video scenes are consistent with unbiased descriptions. These limitations fail to align with real-world scenarios since descriptions can be influenced by annotator biases, diverse writing styles, and varying textual perspectives. To overcome the aforementioned problems, we introduce \texttt{WAVER}, a cross-domain knowledge distillation framework via vision-language models through open-vocabulary knowledge designed to tackle the challenge of handling different writing styles in video descriptions. \texttt{WAVER} capitalizes on the open-vocabulary properties that lie in pre-trained vision-language models and employs an implicit knowledge distillation approach to transfer text-based knowledge from a teacher model to a vision-based student. Empirical studies conducted across four standard benchmark datasets, encompassing various settings, provide compelling evidence that \texttt{WAVER} can achieve state-of-the-art performance in text-video retrieval task while handling writing-style variations. The code is available at: \url{https://github.com/Fsoft-AIC/WAVER}

\end{abstract}

\vspace{-1.3em}
\section{Introduction}
\vspace{-1em}
Text-video retrieval (TVR), the task of retrieving videos based on textual queries, has grown significantly in multimedia information retrieval. Current works focus on cross-modal feature matching, assuming consistent video scenes and unbiased descriptions. However, existing TVR datasets~\cite{DBLP:conf/cvpr/XuMYR16,DBLP:conf/acl/ChenD11,DBLP:conf/iccv/WangWCLWW19,DBLP:conf/iccv/HendricksWSSDR17} are manually annotated by numerous annotators, introducing complexities due to variations in imperfect annotations, writing styles, and diverse perspectives. Consequently, this results in distinct semantic interpretations among descriptions associated with the same video. Moreover, the advent of large-scale pre-trained Vision-Language Models (VLMs) like \texttt{CLIP}~\cite{DBLP:conf/icml/RadfordKHRGASAM21} has marked significant progress in TVR over recent years. However, existing methods fully fine-tune \texttt{CLIP} for feature extraction and fusion in a brute-force manner, missing out on fully harnessing its pre-trained knowledge.

To address the aforementioned challenges, we propose \texttt{WAVER} with a cross-domain knowledge distillation (KD) mechanism to address a novel task referred to as ``writing-style agnostic". \texttt{WAVER}'s primary objective is to alleviate the influence of diverse writing styles on TVR by exploring the open-vocabulary (open-vocab) properties present in VLMs through our cross-domain KD mechanism. Within this method, we first compile a Video Content Dictionary (VCD) that comprises phrases. Each phrase represents a specific activity and is among the top-$k$ relevant activities for a given video. This knowledge source, derived from the VCD and extracted from large-scale datasets (usually training sets, but not limited to) encompasses a multitude of writing styles contributed by numerous annotators. The knowledge from VCD is treated as a teacher, embodying comprehensive information enriched by a wide spectrum of writing styles. Our proposed cross-domain KD aims to distill text-based knowledge from the large-capacity teacher (i.e., comprising various writing styles) into a vision-based student model when presented with a specific video and its associated content. This approach equips \texttt{WAVER} with the flexibility to effectively handle a wide variety of writing styles. To evaluate the effectiveness of our approach, we conducted a series of comprehensive experiments and ablation studies across four prominent benchmarks: \textit{MSR-VTT}~\cite{DBLP:conf/cvpr/XuMYR16},
\textit{MSVD}~\cite{DBLP:conf/acl/ChenD11}, \textit{VATEX}~\cite{DBLP:conf/iccv/WangWCLWW19}, and \textit{DiDeMo}~\cite{DBLP:conf/iccv/HendricksWSSDR17}.

\vspace{-1.3em}
\section{Related Work}
\vspace{-1.0em}
\textbf{Vision-Language Models.} 
VLMs~\cite{DBLP:conf/icml/RadfordKHRGASAM21,DBLP:conf/icml/0001LXH22,DBLP:conf/icml/0008LSH23} have been applied to various vision tasks with the use of open-vocab knowledge. For example, for image classification, \texttt{ALIGN}~\cite{DBLP:conf/icml/JiaYXCPPLSLD21} and \texttt{UniCL}~\cite{DBLP:conf/cvpr/YangLZXLYG22} improve accuracy by matching images with text descriptions; for object detection, \texttt{X-DETR}~\cite{DBLP:conf/eccv/CaiKRBTBS22} and \texttt{OWL-ViT}~\cite{DBLP:conf/eccv/MindererGSNWDMADSWZKH22} utilize VLMs for localization and recognition; for image segmentation, \texttt{DenseCLIP} \cite{DBLP:conf/cvpr/RaoZ0TZH0L22} and \texttt{OpenSeg}~\cite{DBLP:conf/eccv/GhiasiGCL22} utilizes VLMs for pixel-level classification. TVR methods~\cite{DBLP:journals/ijon/LuoJZCLDL22,DBLP:journals/corr/abs-2106-11097,DBLP:conf/cvpr/GeGLLSQL22,DBLP:conf/eccv/Gabeur0AS20,DBLP:conf/cvpr/LeiLZGBB021} in the second group have also benefited from VLMs by extending \texttt{CLIP} to train text-video matching models using a contrastive loss. However, most of these approaches fully fine-tune VLMs, under-utilizing pre-trained multi-modal information in videos. In our work, we leverage both fine-tuning and pre-trained VLMs feature to enhance TVR performance further.\\
\textbf{Knowledge Distillation.} 
The concept of KD~\cite{DBLP:journals/corr/HintonVD15} is to transfer knowledge from a teacher model with robust knowledge to a student model while maintaining high accuracy. Existing KD methods~\cite{DBLP:conf/iccv/TungM19, DBLP:conf/sigmod/0002Z0KGJ23} explicitly maintain teacher and student models to focus mainly on the technique to transfer knowledge between them. In contrast,
to the best of our knowledge, our framework is the first study that uses open-vocab knowledge as the teacher to implicitly transfer the large knowledge from the teacher to the student to address the TVR task.


\begin{figure}[!t]  
    \includegraphics[width=1.0\linewidth]{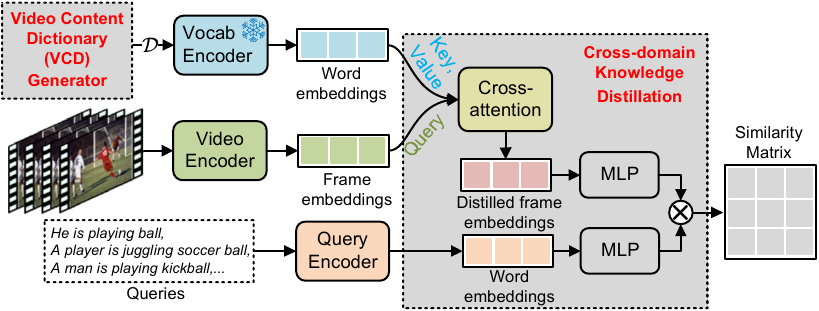}
    \caption{\textbf{Overview framework of \texttt{WAVER}.} We propose a cross-domain KD, in which open-vocab knowledge from pre-trained VLM implicitly acts as the teacher. By distilling text-based knowledge from the large-capacity teacher into a vision-based student model (described in Section~\ref{sec:med}) to handle writing-style variations.}
    \label{fig:overview}
\vspace{-1.5em}
\end{figure}

\vspace{-1.8em}
\section{Methodology}
\label{sec:med}
\vspace{-1em}
In \texttt{WAVER}, we extract features from both the video and the query using the \textit{Video Encoder} and \textit{Query Encoder}, respectively, as described in Section \ref{sec:fea}. To tackle the challenge of writing-style variations, our \texttt{WAVER} creates a VCD, which compiles diverse video descriptions produced by different annotators, capturing a range of writing styles (cf. Section~\ref{sec:VCD}). The knowledge extracted from the VCD serves as the teacher, while the \textit{Video Encoder} functions as the student. Subsequently, we introduce Cross-domain KD (cf. Section~\ref{sec:KD}) to transfer the teacher's text-based knowledge to the student's vision. The overall workflow of \texttt{WAVER} is illustrated in Fig.~\ref{fig:overview}.

\noindent
\textbf{Problem Setup.} We have $L$ trimmed videos $\mathcal{V} = \{\textbf{V}^{(l)}\}_{l=1}^L$
accompanied by a set of corresponding textual descriptions $\mathcal{T} = \{\textbf{T}^{(l)}\}_{l=1}^L$.
The primary goal of this problem is to retrieve video $\mathbf{V}^{(l)}$ based on $\mathbf{T}^{(l)}$.
\vspace{-1.7em}
\subsection{Feature Encoders}
\label{sec:fea}
\vspace{-0.7em}
We use Vision Transformer (\texttt{ViT})~\cite{DBLP:conf/iclr/DosovitskiyB0WZ21} from \texttt{CLIP}~\cite{DBLP:conf/icml/RadfordKHRGASAM21} as our \textit{Video Encoder}.
Given a video $\mathbf{V}^{(i)}$ composed of $N$ frames, we extract visual features $\mathbf{v}^{(i)} = f_{v}(\mathbf{V}^{(i)}|{\theta}_v)$.
Here, ${\mathbf{v}}^{(i)}=\langle{\mathbf{v}}^{(i)}_1, \ldots, {\mathbf{v}}^{(i)}_N\rangle$ is an embedding sequence of $N$ frames and $f_{v}(\cdot|{\theta}_v)$ is the \textit{Video Encoder}, which is parameterized by \texttt{ViT}'s weights ${\theta}_v$.
As for the \textit{Query Encoder} for textual feature extraction, we employ \texttt{CLIP}'s text encoder. 
Given a textual description $\mathbf{T}^{(i)}$, which consists of $S$ tokens, we obtain textual features $\mathbf{t}^{(i)} = f_{t}(\mathbf{T}^{(i)}|{\theta}_t)$.
Here, ${\mathbf{t}}^{(i)}=\langle{\mathbf{t}}^{(i)}_1, \ldots, {\mathbf{t}}^{(i)}_S\rangle$ is an embedding sequence of $S$ tokens and $f_{t}(\cdot|{\theta}_t)$ is the \textit{Query Encoder}, which is parameterized by \texttt{Transformer}'s weights ${\theta}_t$.

\vspace{-1.9em}
\subsection{Video Content Dictionary Generator}
\label{sec:VCD}
\vspace{-0.5em}

\begin{figure}[!t]  
    \includegraphics[width=1.0\linewidth]{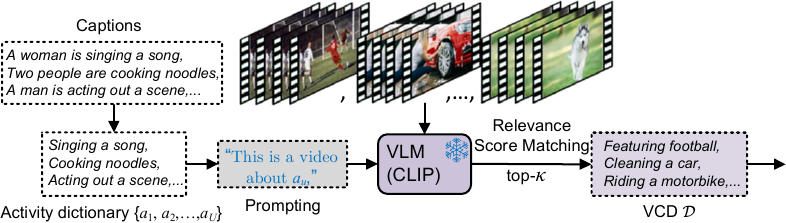}
    \caption{\textbf{The details of Video Content Dictionary Generator.} As described in Section \ref{sec:VCD}, we first form a list of all activities extracted from each caption. Then we propose a matching scheme to form a set of video-related vocabularies and select the top-$\kappa$ most relevant vocabularies to create a VCD.}
    \label{fig:vcd}
\vspace{-1.7em}
\end{figure}

To explore the open-vocab capability in a pre-trained VLM, we propose to construct a VCD using auxiliary information from the VLM to identify the most relevant knowledge. 
As depicted in Fig.\ref{fig:vcd}, given a TVR dataset, we first use \textit{spaCy}\footnote{https://spacy.io} to extract existing activities (i.e., verb phrases) in each caption. As a result, we form a list of all activities $\{a_1, a_2, ..., a_U\}$, where $U$ is the number of activities. We then propose a matching strategy to form a set of video-related vocabularies. For each activity $a_u$, we add a manually-designed prompt to a prefix to form a full query i.e., $q_u=$ ``\texttt{This is a video about} $a_{u}$". These queries $q_i$ are processed through a frozen pre-trained \texttt{CLIP}'s text encoder, resulting in embedding vocabularies ($\mathbf{h}_1, \ldots, \mathbf{h}_U$). Additionally, we use a frozen pre-trained \texttt{CLIP}'s visual encoder to extract frame-level embedding ($\mathbf{q}_1^{(i)}, \ldots, \mathbf{q}_N^{(i)}$) for each video $\mathbf{V}^{(i)}$. The global video embedding ($\mathbf{e}^{(i)}$) for each video is obtained by Eq.~\ref{eq:e}.
\begin{align}
    \mathbf{e}^{(i)}=\frac{1}{N}\sum_{t=1}^N\mathbf{q}^{(i)}_t.
\label{eq:e}
\end{align}
Next, we use function $s$ to measure the similarity between each video and the entire vocabulary set by calculating the cosine similarity between the video's global embedding $\mathbf{e}^{(i)}$ and vocabularies embedding $\mathbf{h}_u$.
\begin{align}
s(\mathbf{e}^{(i)}, \mathbf{h}_u) = \frac{\mathbf{e}^{(i)} \cdot \mathbf{h}_u}
{||{\mathbf{e}}^{(i)}|| \cdot ||{\mathbf{h}}_u||},
\end{align}

\noindent where $\cdot$ denotes the dot product operation, and $\left\| \right\|$ denote the $\ell_2$-norm of the feature vectors.

We then select the top-$\kappa$ most relevant vocabularies \ $d^{(i)} = \langle d^{(i)}_{1}, d^{(i)}_{2}, \ldots, d^{(i)}_{K} \rangle = \text{top-}\kappa(\{s(\mathbf{e}^{(i)}, \mathbf{h}_u)\}_{u=1}^U)$ for each video $\mathbf{V}^{(i)}$. Then, $d^{(i)}$ serves as a vocabulary. As a result, we finally form a VCD denoted as $\mathcal{D} = \{d^{(i)}\}_{i=1}^{L}$.

\vspace{-1.5em}
\subsection{Cross-domain Knowledge Distillation}
\vspace{-0.5em}
\label{sec:KD}

By utilizing VCD $\mathcal{D}$, our \texttt{WAVER} framework efficiently highlights the open-vocab property in pre-trained VLM to transfer the foundational knowledge from teacher $\mathcal{D}$ to the video encoder. 
As illustrated in Fig.~\ref{fig:overview}, we leverage the knowledge extracted from the VCD $\mathcal{D}$ as the teacher, while the \textit{Video Encoder} functions as the student. To facilitate the transfer of knowledge from the teacher to the student, we employ cross-attention mechanism~\cite{DBLP:conf/iccv/ChenFP21}. We initiate the process by creating prompts $\mathbf{C}^{(i)}$ of the form ``\texttt{This is a video about} $d^{(i)}$" for each vocabulary $d^{(i)}$. These prompts, $\mathbf{C}^{(i)}$, are then inputted into the \textit{Vocab Encoder}, $f_{c}(\cdot|\theta_{c})$, which is a frozen pre-trained \texttt{CLIP}'s text encoder. The output $\mathbf{c}^{(i)}$ is computed as $\mathbf{c}^{(i)} = f_{c}(\mathbf{C}^{(i)}|{\theta}_{c})$, where $\theta_{c}$ is network weights. As a result, we acquire a general knowledge corpus $\mathcal{C}$ independent of any specific writing style, and it is defined as $\mathcal{C} = \{\mathbf{c}^{(i)}\}_{i=1}^L$.

Given a video $\mathbf{V}^{(i)}$ containing visual features ${\mathbf{v}}^{(i)}=\langle{\mathbf{v}}^{(i)}_1, \ldots, {\mathbf{v}}^{(i)}_N\rangle$, we define the cross-attention between ${\mathbf{v}}^{(i)}$ and $\mathcal{C}$ as outlined in Eq. \ref{eq:ca}, where ${\mathbf{v}}^{(i)}$ takes on the role of the query, and $\mathcal{C}$ serves as the key/value. As a result, we obtain a distilled video embedding $\thickbar{\mathbf{v}}^{(i)}$ at the student side.
\begin{align}
\label{eq:ca}
\thickbar{\mathbf{v}}^{(i)}=\operatorname{softmax}(\frac{{\mathbf{v}}^{(i)}\mathcal{C}^\top}{\sqrt{z}}) \mathcal{C},
\end{align}

\noindent where $z$ is the scaling factor.
\vspace{-1.35em}
\subsection{Learning Objective Functions}
\vspace{-0.5em}
Inspired by~\cite{DRLTVR2022}, we align the distilled video embedding $\thickbar{\mathbf{v}}^{(i)}$ and query embedding $\mathbf{t}^{(i)}$ with two Multi-layer Perceptrons (MLPs), respectively.
The MLPs project $\thickbar{\mathbf{v}}^{(i)}$ and $\mathbf{t}^{(i)}$ into a normalized, lower-dimensional representation in the shared latent space.
Then, we calculate the similarity 
$s(\mathbf{t}^{(i)}, \thickbar{\mathbf{v}}^{(i)})$, where $f_\phi(\cdot|\theta_\phi)$ and $f_\psi(\cdot|\theta_\psi)$ are MLPs parameterized by a stack of three Fully-Connected (FC) layers  $\theta_\phi$ and $\theta_\psi$, respectively.
Next, during the training phase, we aim to pull the query embedding $\mathbf{t}^{(i)}$ and the distilled video embedding $\thickbar{\mathbf{v}}^{(i)}$ when they are related and push them apart when they are not related. 
To achieve this, we employ the $\mathrm{InfoNCE}$ loss~\cite{DBLP:journals/corr/abs-1807-03748} to maximize the similarity for matching pairs ${\mathbf{t}^{(i)}, \thickbar{\mathbf{v}}^{(i)}}$ and minimize it for other pairs. This loss function is used for both the Video-to-Text and Text-to-Video problems.

\begin{align}
\mathcal{L}_{t \xrightarrow{} v} &=-\frac{1}{B} \sum_{i}^{B} \log \frac{\exp \left(s\left(\mathbf{t}^{(i)}, \thickbar{\mathbf{v}}^{(i)}\right)\right/\tau)}{\sum_{j=1}^{B} \exp \left(s\left(\mathbf{t}^{(i)}, \thickbar{\mathbf{v}}^{(j)}\right)\right/\tau)},
\end{align}

\begin{align}
\mathcal{L}_{v \xrightarrow{} t} &=-\frac{1}{B} \sum_{i}^{B} \log \frac{\exp \left(s\left(\thickbar{\mathbf{v}}^{(i)}, \mathbf{t}^{(i)}\right)\right/\tau)}{\sum_{j=1}^{B} \exp \left(s\left(\thickbar{\mathbf{v}}^{(i)}, \mathbf{t}^{(j)}\right)\right/\tau)},
\end{align}

\noindent
where $\tau$ is a learnable temperature parameter and $B$ is the batch size.The overall $\mathrm{InfoNCE}$ loss is computed as:
\begin{align}
\mathcal{L}_{\mathrm{InfoNCE}} &=\frac{1}{2}(\mathcal{L}_{t \xrightarrow{} v} + \mathcal{L}_{v \xrightarrow{} t}).
\end{align}

\begin{table}
\footnotesize
\centering
\setlength{\tabcolsep}{3pt}
\renewcommand{\arraystretch}{0.8}
\vspace{-1em}
\caption{\textbf{T2V} comparison between our \texttt{WAVER} with existing SOTA methods on \textit{MSR-VTT}. 
}
\vspace{-1em}
\label{table:msrvtt}
\resizebox{0.9\columnwidth}{!}{%
\begin{tabular}{lc|ccccc}
\toprule
\textbf{Type} & \textbf{Method} & \textbf{R@1$\uparrow$} & \textbf{R@5$\uparrow$} & \textbf{R@10$\uparrow$} & \textbf{MdR$\downarrow$} & \textbf{MnR$\downarrow$}\\
\midrule
w/o. \texttt{CLIP} & \texttt{CE}~\cite{DBLP:conf/bmvc/LiuANZ19} & 20.9 & 48.8 & 62.4 & 6.0 & 28.2 \\
w/o. \texttt{CLIP} & \texttt{ClipBERT}~\cite{DBLP:conf/cvpr/LeiLZGBB021} & 22.0 & 46.8 & 59.9 & 6.0 & - \\
w/o. \texttt{CLIP} & \texttt{MMT}~\cite{DBLP:conf/eccv/Gabeur0AS20} & 26.6 & 57.1 & 69.6 & 4.0 & - \\

w/o. \texttt{CLIP} & \texttt{SupportSet}~\cite{DBLP:conf/iclr/Patrick0AMHHV21}  & 30.1 & 58.5 & 69.3 & 3.0 & - \\
w/o. \texttt{CLIP} & \texttt{Frozen}~\cite{DBLP:conf/iccv/BainNVZ21} & 32.5 & 61.5 & 71.2 & 3.0 & - \\
w/o. \texttt{CLIP} & \texttt{BridgeFormer}~\cite{DBLP:conf/cvpr/GeGLLSQL22} & 37.6 & 64.8 & 75.1 & - & - \\
w/o. \texttt{CLIP} & \texttt{TMVM}~\cite{DBLP:conf/nips/LinW0ZG0S22} & 36.2 & 64.2 & 75.7 & 3.0 & - \\

w/o. \texttt{CLIP} & \texttt{Clover}~\cite{DBLP:conf/cvpr/HuangLFWSJ23} & 40.5 & 69.8 & 79.4 & 2.0 & - \\
\midrule
\texttt{ViT-B/32} & \texttt{CenterCLIP}~\cite{DBLP:conf/sigir/ZhaoZWY22} & 44.2 & 71.6 & 82.1 & 2.0 & 15.1 \\
\texttt{ViT-B/32} & \texttt{CLIP4Clip}~\cite{DBLP:journals/ijon/LuoJZCLDL22} & 44.5 & 71.4 & 81.6 & 2.0 & 15.3 \\
\texttt{ViT-B/32} & \texttt{VoP}~\cite{DBLP:conf/cvpr/HuangGPJLLW23} & 44.6 & 69.9 & 80.3 & 2.0 & 16.3 \\
\texttt{ViT-B/32} & \texttt{CAMoE}~\cite{DBLP:journals/corr/abs-2109-04290} & 44.6 & 72.6 & 81.8 & 2.0 & 13.3 \\
\texttt{ViT-B/32} & \texttt{CLIP2Video}~\cite{DBLP:journals/corr/abs-2106-11097} & 45.6 & 72.6 & 81.7 & 2.0 & 14.6 \\
\texttt{ViT-B/32} & \texttt{X-Pool}~\cite{DBLP:conf/cvpr/GortiVMGVGY22} & 46.9 & 72.8 & 82.2 & 2.0 & 14.3 \\
\texttt{ViT-B/32} & \texttt{TS2-Net}~\cite{DBLP:conf/eccv/LiuXXCJ22} & 47.0 & 74.5 & 83.8 & 2.0 & 13.0 \\
\texttt{ViT-B/32} & \textbf{\texttt{WAVER} (ours)} & \textbf{47.8} & \textbf{74.6} & \textbf{83.9} & \textbf{2.0} & \textbf{12.8} \\
\midrule
\texttt{ViT-B/16} & \texttt{CLIP2TV}~\cite{DBLP:journals/corr/abs-2111-05610} & 48.3 & 74.6 & 82.8 & 2.0 & 14.9 \\
\texttt{ViT-B/16} & \texttt{CenterCLIP}~\cite{DBLP:conf/eccv/LiuXXCJ22} & 48.4 & 73.8 & 82.0 & 2.0 & 13.8 \\
\texttt{ViT-B/16} & \texttt{TS2-Net}~\cite{DBLP:conf/eccv/LiuXXCJ22} & 49.4 & 75.6 & 85.3 & 2.0 & 13.5 \\
\texttt{ViT-B/16} & \textbf{\texttt{WAVER} (ours)} & \textbf{50.4} & \textbf{77.2} & \textbf{86.4} & \textbf{1.0} & \textbf{10.8} \\
\bottomrule
\end{tabular}}
\vspace{-1.2em}
\end{table}

\begin{table}[thb]
\footnotesize
\centering
\setlength{\tabcolsep}{3pt}
\renewcommand{\arraystretch}{0.8}
\caption{\textbf{T2V} comparison between our \texttt{WAVER} with existing SOTA methods on \textit{MSVD}.}
\vspace{-1em}
\label{table:msvd}
\resizebox{0.9\columnwidth}{!}{%
\begin{tabular}{lc|ccccc}
\toprule
\textbf{Type} & \textbf{Method} & \textbf{R@1$\uparrow$} & \textbf{R@5$\uparrow$} & \textbf{R@10$\uparrow$} & \textbf{MdR$\downarrow$} & \textbf{MnR$\downarrow$}\\
\midrule 
w/o. \texttt{CLIP} & \texttt{CE}~\cite{DBLP:conf/bmvc/LiuANZ19} & 19.8 & 49.0 & 63.8 & 6.0 & - \\
w/o. \texttt{CLIP} & \texttt{SupportSet}~\cite{DBLP:conf/iclr/Patrick0AMHHV21} & 28.4 & 60.0 & 72.9 & 4.0 & - \\
w/o. \texttt{CLIP} & \texttt{Frozen}~\cite{DBLP:conf/iccv/BainNVZ21} & 33.7 & 64.7 & 76.3 & 3.0 & - \\
w/o. \texttt{CLIP} & \texttt{TMVM}~\cite{DBLP:conf/nips/LinW0ZG0S22} & 36.7 & 67.4 & 81.3 & 2.5 & - \\
\midrule
\texttt{ViT-B/16} & \texttt{CLIP4Clip}~\cite{DBLP:journals/ijon/LuoJZCLDL22} & 45.2 & 75.5 & 84.3 & 2.0 & 10.3 \\
\texttt{ViT-B/16} & \texttt{X-Pool}~\cite{DBLP:conf/cvpr/GortiVMGVGY22} & 47.2 & 77.4 & 86.0 & 2.0 & 9.3 \\
\texttt{ViT-B/16} & \textbf{\texttt{WAVER} (ours)} & \textbf{50.2} & \textbf{83.5} & \textbf{88.1} & \textbf{2.0} & \textbf{8.9} \\
\bottomrule
\end{tabular}}
\vspace{-2em}
\end{table}

\begin{table}[thb]
\footnotesize
\setlength{\tabcolsep}{3pt}
\renewcommand{\arraystretch}{0.8}
\vspace{-1.8em}
\caption{\textbf{T2V} comparison between our \texttt{WAVER} with existing SOTA methods on \textit{VATEX}.}
\vspace{-1em}
\label{table:vatex}
\resizebox{\columnwidth}{!}{%
\begin{tabular}{lp{2.3cm}|ccccc}
\toprule
\textbf{Type} & \textbf{Method} & \textbf{R@1$\uparrow$} & \textbf{R@5$\uparrow$} & \textbf{R@10$\uparrow$} & \textbf{MdR$\downarrow$} & \textbf{MnR$\downarrow$}\\
\midrule 
w/o. \texttt{CLIP} & \texttt{HGR}~\cite{DBLP:conf/cvpr/ChenZJW20} & 35.1 & 73.5 & 83.5 & 2.0 & - \\
w/o. \texttt{CLIP} & \texttt{SupportSet}~\cite{DBLP:conf/iclr/Patrick0AMHHV21} & 44.9 & 82.1 & 89.7 & 1.0 & - \\
\midrule
\texttt{ViT-B/16} & \texttt{CLIP4Clip}~\cite{DBLP:journals/ijon/LuoJZCLDL22} & 55.9 & 89.2 & 95.0 & 1.0 & 3.9 \\
\texttt{ViT-B/16} & \texttt{CLIP2Video}~\cite{DBLP:journals/corr/abs-2106-11097} & 57.3 & 90.0 & 95.5 & 1.0 & 3.6\\
\texttt{ViT-B/16} & \texttt{QB-Norm}~\cite{DBLP:conf/cvpr/BogolinCJLA22} & 58.8 & 88.3 & 93.8 & 1.0 & - \\
\texttt{ViT-B/16} & \texttt{TS2-Net}~\cite{DBLP:conf/eccv/LiuXXCJ22} & 59.1 & 90.0 & 95.2 & 1.0 & 3.5 \\
\texttt{ViT-B/16} & \textbf{\texttt{WAVER} (ours)} & \textbf{66.5} & \textbf{93.3} & \textbf{97.0} & \textbf{1.0} & \textbf{2.8} \\
\bottomrule
\end{tabular}}
\vspace{-1.5em}
\end{table}

\begin{table}[thb]
\footnotesize
\centering
\setlength{\tabcolsep}{3pt}
\renewcommand{\arraystretch}{0.8}
\vspace{-1.7em}
\caption {\textbf{T2V} comparison between our \texttt{WAVER} with existing SOTA on \textit{DiDeMo}.}
\label{table:didemo}
\vspace{-1em}
\resizebox{0.8\columnwidth}{!}{%
\begin{tabular}{lc|ccccc}
\toprule
\textbf{Type} & \textbf{Method} & \textbf{R@1$\uparrow$} & \textbf{R@5$\uparrow$} & \textbf{R@10$\uparrow$} & \textbf{MdR$\downarrow$} & \textbf{MnR$\downarrow$}\\
\midrule 

w/o. \texttt{CLIP} & \texttt{CE}~\cite{DBLP:conf/bmvc/LiuANZ19} & 15.6 & 40.9 & - & 8.2 & -\\
w/o. \texttt{CLIP} & \texttt{ClipBERT}~\cite{DBLP:conf/iclr/Patrick0AMHHV21} & 21.1 & 47.3 & 61.1 & 6.3 & -\\
w/o. \texttt{CLIP} & \texttt{Frozen}~\cite{DBLP:conf/iccv/BainNVZ21} & 31.0 & 59.8 & 72.4 & 3.0 & - \\
w/o. \texttt{CLIP} & \texttt{TMVM}~\cite{DBLP:conf/nips/LinW0ZG0S22} & 36.5 & 64.9 & 75.4 & 3.0 & - \\
\midrule
\texttt{ViT-B/16} & \texttt{CLIP4Clip}~\cite{DBLP:journals/ijon/LuoJZCLDL22} & 42.8 & 68.5 & 79.2 & 2.0 & 18.9 \\
\texttt{ViT-B/16} & \texttt{TS2-Net}~\cite{DBLP:conf/eccv/LiuXXCJ22} & 41.8 & 71.6 & 82.0 & 2.0 & 14.8 \\
\texttt{ViT-B/16} & \texttt{HunYuan}~\cite{Tencenttvr} & 45.0 & 75.6 & 83.4 & 2.0 & 12.0 \\
\texttt{ViT-B/16} & \textbf{\texttt{WAVER} (ours)} & \textbf{49.2} & \textbf{77.2} & \textbf{85.6} & \textbf{2.0} & \textbf{11.2} \\
\bottomrule
\end{tabular}}
\vspace{-1.9em}
\end{table}

\vspace{-1.3em}
\section{Experiments}
\vspace{-1em}
\subsection{Datasets, Metrics \& Implementation Details}
\vspace{-0.5em}
We benchmark our \texttt{WAVER} on Text-to-Video (T2V) task on \textit{MSR-VTT}~\cite{DBLP:conf/cvpr/XuMYR16}, \textit{MSVD}~\cite{DBLP:conf/acl/ChenD11}, \textit{VATEX}~\cite{DBLP:conf/iccv/WangWCLWW19}, and  \textit{DiDeMo}~\cite{DBLP:conf/iccv/HendricksWSSDR17}. \textit{MSR-VTT} contains 10,000 videos, 20 descriptions per video. We train on 9,000 videos and test on 1,000 videos. 
\textit{MSVD} contains 1,970 videos with multiple descriptions in various languages. We use 1,200 videos for training and 670 for testing, considering only English descriptions.
\textit{VATEX} contains over 40,000 videos and 825,000 captions in both English and Chinese. We only consider English descriptions in our experiment with 26,000/1,500/1,500 videos for training/validation/testing. \textit{DiDeMo} contains 10,000 videos with over 40,000 text descriptions. Training on 9,000 videos, we report results on the remaining 1,000 videos.

Following previous works~\cite{DBLP:journals/ijon/LuoJZCLDL22}, we report the result of the testing set with evaluation on multiple captions per video, except for \textit{MSR-VTT}, where each video has only one caption. We evaluate the performance on the T2V task with various metrics including \textit{R@}1, \textit{R@}5, \textit{R@}10, MdR, and MnR.

We set the token length to 32, the video sample frame to 12 for \textit{MSR-VTT} and \textit{MSVD} and the token length to 64, the video sample frame to 64 for \textit{DiDeMo} and \textit{VATEX}. The scaling factor is set $z$ to 64 and the batch size is 126. and the initial learning rate is set to $10^{-4}$ and $10^{-7}$  for non-/ \texttt{CLIP}-based methods, respectively. The number of epochs is 5 for both versions. The model is implemented in PyTorch~\cite{paszke2019pytorch} and trained by 2 $\times$ A100 GPUs. 
\vspace{-1.3em}
\subsection{Comparison with State-of-the-art Methods}
\vspace{-0.5em}
In \textit{Table}~\ref{table:msrvtt}, our \texttt{WAVER} with both \texttt{CLIP}'s video encoder backbones versions (i.e., \texttt{ViT-B/32}, \texttt{ViT-B/16}) achieves SOTA results.
With \texttt{ViT-B/32}, we achieve 47.8 \textit{R@}1 surpassing the runner-up \texttt{TS2-Net}~\cite{DBLP:conf/eccv/LiuXXCJ22} by 0.8 point in T2V. 
With \texttt{ViT-B/16}, \texttt{WAVER} achieves 50.4 \textit{R@}1 outperforming the runner-up \texttt{TS2-Net}~\cite{DBLP:conf/eccv/LiuXXCJ22} by 1.0 point in T2V. 
\textit{Table}~\ref{table:msvd} shows that we \texttt{WAVER} significantly outperforms the runner-up \texttt{X-Pool}~\cite{DBLP:conf/cvpr/GortiVMGVGY22} by 3.0 points \textit{R@}1 achieving SOTA performance of 50.2 \textit{R@}1.
\textit{Table}~\ref{table:vatex} and
\textit{Table}~\ref{table:didemo} further demonstrate that \texttt{WAVER} achieves SOTA performance on both \textit{VATEX} and \textit{DiDeMo} datasets. For \textit{VATEX}, our \texttt{WAVER} exhibits outstanding performance with a 66.5\% \textit{R@}1 score, surpassing the runner-up \texttt{TS2-Net}~\cite{DBLP:conf/eccv/LiuXXCJ22} and \texttt{QB-Norm}~\cite{DBLP:conf/cvpr/BogolinCJLA22} 7.4 and 7.7 points on \textit{R@}1. Additionally, for \textit{DiDeMo}, our approach outperforms the recent SOTA \texttt{X-Pool}~\cite{DBLP:conf/cvpr/GortiVMGVGY22} with 2.0 points on \textit{R@}1. 
In smaller-scale datasets like \textit{MSVD}, we emphasize the effectiveness of our proposed KD during transferring the knowledge from the general teacher $\mathcal{C}$ to the student video encoder function, enhancing distilled video feature $\thickbar{\mathbf{v}}$.
\vspace{-0.9em}

\begin{table}[thb]
\centering
\setlength{\tabcolsep}{7pt}
\renewcommand{\arraystretch}{0.7}
\caption{Effectiveness of Cross-domain KD on \textit{MSR-VTT}.}
\label{table:ablation_msrvtt}
\vspace{-1em}
\resizebox{0.7\columnwidth}{!}{%
\begin{tabular}{l|cccc}
\toprule
\textbf{Method} & \textbf{R@1$\uparrow$} & \textbf{R@5$\uparrow$} & \textbf{R@10$\uparrow$} & \textbf{MdR$\downarrow$} \\
\midrule 
Baseline & 45.6 & 72.8 & 82.2 & 2.0 \\
\toprule
\multicolumn{5}{c}{ \centering \textbf{+ Different values of top-$\kappa$} }\\
$\kappa=1$ & 47.6 & 73.3 & 82.4 & 2.0 \\
$\kappa=3$ & 47.7 & 73.1 & 82.8 & 2.0 \\
\textbf{$\kappa=5$} & \textbf{47.8} & \textbf{74.6} & \textbf{83.9} & \textbf{2.0} \\
$\kappa=7$ & 47.4 & 73.3 & 83.8 & 2.0 \\
$\kappa=9$ & 46.9 & 73.5 & 84.8 & 2.0 \\
\bottomrule
\end{tabular}}
\vspace{-1.2em}
\end{table}

\begin{table}[thb]
\centering
\setlength{\tabcolsep}{6pt}
\renewcommand{\arraystretch}{0.8}
\vspace{-0.7em}
\caption{Performance of \texttt{WAVER} on \textit{MSR-VTT} testing set using  various VCD $\mathcal{D}$ ( \textit{MSR-VTT}, \textit{MSVD}, \textit{VATEX}, \textit{DiDeMo} ).}
\label{table:vocab_dataset}
\vspace{-1em}
\resizebox{0.8\columnwidth}{!}{%
\begin{tabular}{ll|ccccc}
\toprule
\textbf{Vocab} & \textbf{No. Vocab} &
\textbf{R@1$\uparrow$} & \textbf{R@5$\uparrow$} & \textbf{R@10$\uparrow$} & \textbf{MdR$\downarrow$} \\
\midrule 
\textit{MSR-VTT} & 50,482 & \textbf{47.8} & \textbf{74.6} & \textbf{83.9} & \textbf{2.0} \\
\textit{MSVD} & 21,168 & 47.1 & 74.0 & 83.8 & 2.0  \\
\textit{VATEX} & 108,596 & 47.3 & 73.6 & 83.7 & 2.0 \\
\textit{DiDeMo} & 9,132 & 47.6 & 72.7 & 82.1 & 2.0  \\
\bottomrule
\end{tabular}}
\vspace{-2.3em}
\end{table}

\begin{table}[thb]
\centering
\setlength{\tabcolsep}{8pt}
\renewcommand{\arraystretch}{0.7}
\vspace{-1.5em}
\caption{Evaluation on \textit{MSR-VTT} testing set with different writing styles randomly selected using various seed values.}
\label{table:single-caption}
\vspace{-1em}
\resizebox{0.8\columnwidth}{!}{%
\begin{tabular}{ll|cccc}
\toprule
\textbf{Method} & \textbf{Seed} & 
\textbf{R@1$\uparrow$} & \textbf{R@5$\uparrow$} & \textbf{R@10$\uparrow$}& \textbf{MdR$\downarrow$}\\
\midrule 
Baseline & 16 & 45.0 & 72.4 & 81.2 & 2.0 \\
Baseline & 171 & 45.2 & 72.5 & 81.5 & 2.0 \\
Baseline & 1710 & 44.6 & 71.8 & 81.0 & 2.0 \\
Baseline & 2804 & 44.4 & 71.6 & 80.8 & 2.0\\
\hline
\textit{Mean} & - & 44.8 & 72.1 & 81.1 & 2.0 \\
\textit{Std} & - & 0.37 & 0.44 &  0.30 & 0.0 \\
\toprule 
\texttt{WAVER} & 16 & 47.3 & 74.3 & 83.5 & 2.0 \\
\texttt{WAVER} & 171 & 47.5 & 74.5 & 83.8 & 2.0 \\
\texttt{WAVER} & 1710 & 47.2 & 74.2 & 83.4 & 2.0 \\
\texttt{WAVER} & 2804 & 47.1 & 74.1 & 83.3 & 2.0 \\
\hline
\textit{Mean} & - & \textbf{47.3} & \textbf{74.3} & \textbf{83.5} & \textbf{2.0} \\
\textit{Std} & - & \textbf{0.27} &  \textbf{0.17} & \textbf{0.22} & \textbf{0.0} \\
\bottomrule
\end{tabular}}
\end{table}

\begin{table}[thb]
\centering
\setlength{\tabcolsep}{7pt}
\renewcommand{\arraystretch}{0.8}
\vspace{-0.7em}
\caption{Performance of \texttt{WAVER} on 100 videos from \textit{MSR-VTT} testing set conducted by four different annotators.}
\label{table:100_video}
\vspace{-1em}
\resizebox{0.8\columnwidth}{!}{%
\begin{tabular}{lc|cccc}
\toprule
\textbf{Method} & \textbf{Annotators} &
\textbf{R@1$\uparrow$} & \textbf{R@5$\uparrow$} & \textbf{R@10$\uparrow$}& \textbf{MdR$\downarrow$}\\
\midrule 
Baseline & \#1 & 67.3 & 91.4 & 94.1 & 1.0 \\
Baseline & \#2 & 65.7 & 90.2 & 93.3 & 1.0 \\
Baseline & \#3 & 66.5 & 90.7 & 93.8 & 1.0 \\
Baseline & \#4 & 64.8 & 89.9 & 92.2 & 1.0 \\
\toprule 
\texttt{WAVER} & \#1 & 75.0 & 95.2 & 97.4 & 1.0 \\
\texttt{WAVER} & \#2 & 74.2 & 94.5 & 96.9 & 1.0\\
\texttt{WAVER} & \#3 & 74.4 & 94.9 & 97.2 & 1.0 \\
\texttt{WAVER} & \#4 & 73.8 & 94.0 & 96.5 & 1.0 \\
\bottomrule
\end{tabular}}
\vspace{-1.8em}
\end{table}

\subsection{Ablation Study}
\vspace{-0.5em}
\textbf{Effectiveness of Cross-domain KD.} In \textit{Table}~\ref{table:ablation_msrvtt}, we highlight the impact of cross-domain KD in \texttt{WAVER} by comparing it with the baseline model, \texttt{Clip4clip}~\cite{DBLP:journals/ijon/LuoJZCLDL22}. Specifically, we observe that when we disable the \textit{Cross-domain KD} module, the baseline framework essentially functions as a basic TVR model, resulting in modest performance. This underscores the effectiveness of our KD method in enhancing accuracy and bolstering model robustness. Moreover, we experiment with different values of $\kappa$ ranging from 1, 3, 5, 7, to 9, to assess their impact on performance. We note that when $\kappa$ is excessively large, the framework encounters difficulty distilling discriminative features from the captions, likely due to the captions being over-specific. Conversely, when $\kappa$ is too small, the captions may fail to fully leverage the open-vocab knowledge embedded in the pre-trained VLM, as they lack sufficient semantic context, ultimately leading to sub-optimal results. We attain the most favorable outcomes when $\kappa = 5$.\\
\textbf{Robustness of \texttt{WAVER}.}
We investigate the robustness of the \texttt{WAVER} model by employing different vocabulary datasets to construct VCD $\mathcal{D}$.  In \textit{Table}~\ref{table:vocab_dataset}, we present the performance of MSR-VTT, where the VCD $\mathcal{D}$ is generated from various datasets \textit{MSR-VTT}, \textit{MSVD}, \textit{VATEX}, and \textit{DiDeMo}. Even when the VCD $\mathcal{D}$ is created using different datasets, the framework consistently achieves high accuracy, with negligible differences in the results. This underscores the robustness of the open-vocab knowledge embedded within \texttt{WAVER} framework.\\
\textbf{\texttt{WAVER} in \textit{Writing-style Agnostic} Task.} 
In \textit{Table}~\ref{table:single-caption}, in addition to the best-performing as in the default setting, we also introduce writing-style diversity by selecting a random writing style for each video evaluation. This randomness is achieved using a random seed, and a random writing style is represented by a randomly selected caption. \textit{Table}~\ref{table:single-caption} demonstrates \texttt{WAVER}'s capacity to retrieve the target videos consistently and accurately, regardless of writing style. Compared to the baseline, the standard deviation (Std) of \texttt{WAVER} highlights its remarkable consistency in retrieving the target videos, even when the writing style of each video's description is altered. To further illustrate \texttt{WAVER}'s effectiveness in handling diverse writing styles, we engaged four annotators to evaluate 100 videos from the \textit{MSR-VTT} testing set. It's important to note that each video contains 20 captions. Based on their writing style, each annotator selected one caption out of the 20 for each video. \textit{Table}~\ref{table:100_video} underscores that despite the biases introduced by different annotators in choosing corresponding captions for each video, our \texttt{WAVER}' results outperform the baseline approach consistently. This study introduces a promising avenue for future research within the TVR task.

\vspace{-1.3em}
\section{Conclusion \& discussion}
\vspace{-1em}
In this work, we have presented \texttt{WAVER}, a writing-style agnostic video retrieval via distilling vision-language models through open-vocab knowledge framework. Our \texttt{WAVER} is a novel framework for the TVR task, where we proposed a cross-domain knowledge distillation mechanism through open-vocabulary properties to effectively utilize the powerful representation knowledge from the pre-trained VLM. To further highlight the applicability of the open-vocabulary properties in dealing with different semantic meanings, we denote a new task namely \textit{writing-style agnostic} task, which evaluates the consistency of the retrieval results from different query descriptions. We hope that our work will inspire future research of \textit{writing-style agnostic} problem and the potential of TVR task in addressing this problem.

\clearpage
\newpage
{\footnotesize
\bibliographystyle{IEEEbib}
\bibliography{main}
}
\end{document}